%% file: main.tex
\documentclass{article}
\usepackage{iclr2027_conference,times}

\iclrfinalcopy  

\input{math_commands.tex}

\usepackage{amsmath,amssymb,mathtools}
\usepackage{booktabs}
\usepackage{graphicx}
\usepackage{enumitem}
\usepackage{wrapfig}
\usepackage{makecell}
\usepackage{multirow}
\usepackage{array}
\usepackage{xcolor}
\usepackage{colortbl}
\definecolor{First}{rgb}{0.95, 0.62, 0.61}
\definecolor{Second}{rgb}{0.97,0.81,0.63}
\usepackage[hidelinks]{hyperref}
\usepackage{url}
\usepackage{caption}
\input{macros.tex}

\title{\method{}: Learning Executable Humanoid Motions Directly from Monocular Video}

\author{
\textbf{Tianyu Xiong\textsuperscript{1,*} \quad
Yi Lu\textsuperscript{1,*} \quad
Jinrui Wang\textsuperscript{1} \quad
Ziqi Liang\textsuperscript{1} \quad
Dandan Lei\textsuperscript{3} \quad
Xiaoyang Zhou\textsuperscript{4}} \\
\textbf{Xiao-xiao Long\textsuperscript{2} \quad
Qiu Shen\textsuperscript{1,\ensuremath{\dagger}} \quad
Xun Cao\textsuperscript{1}} \\[0.7em]
\textsuperscript{1}School of Electronic Science and Engineering,
Nanjing University, Nanjing, China \\
\textsuperscript{2}School of Intelligence Science and Technology,
Nanjing University, Suzhou, China \\
\textsuperscript{3}Jiangsu Mobile Information System Integration Co., Ltd.,
Nanjing, China \\
\textsuperscript{4}China Mobile Zijin (Jiangsu) Innovation Research Institute Co., Ltd.,
Nanjing, China 
}

\begin{document}
\maketitle
\vspace{-5mm}

\begingroup
\renewcommand{\thefootnote}{\fnsymbol{footnote}}
\footnotetext[1]{Equal contribution.\qquad
\textsuperscript{\ensuremath{\dagger}}Corresponding author.}
\endgroup

\noindent\begin{minipage}{\linewidth}
  \captionsetup{type=figure,hypcap=false}
  \centering
  \includegraphics[width=0.9\linewidth]{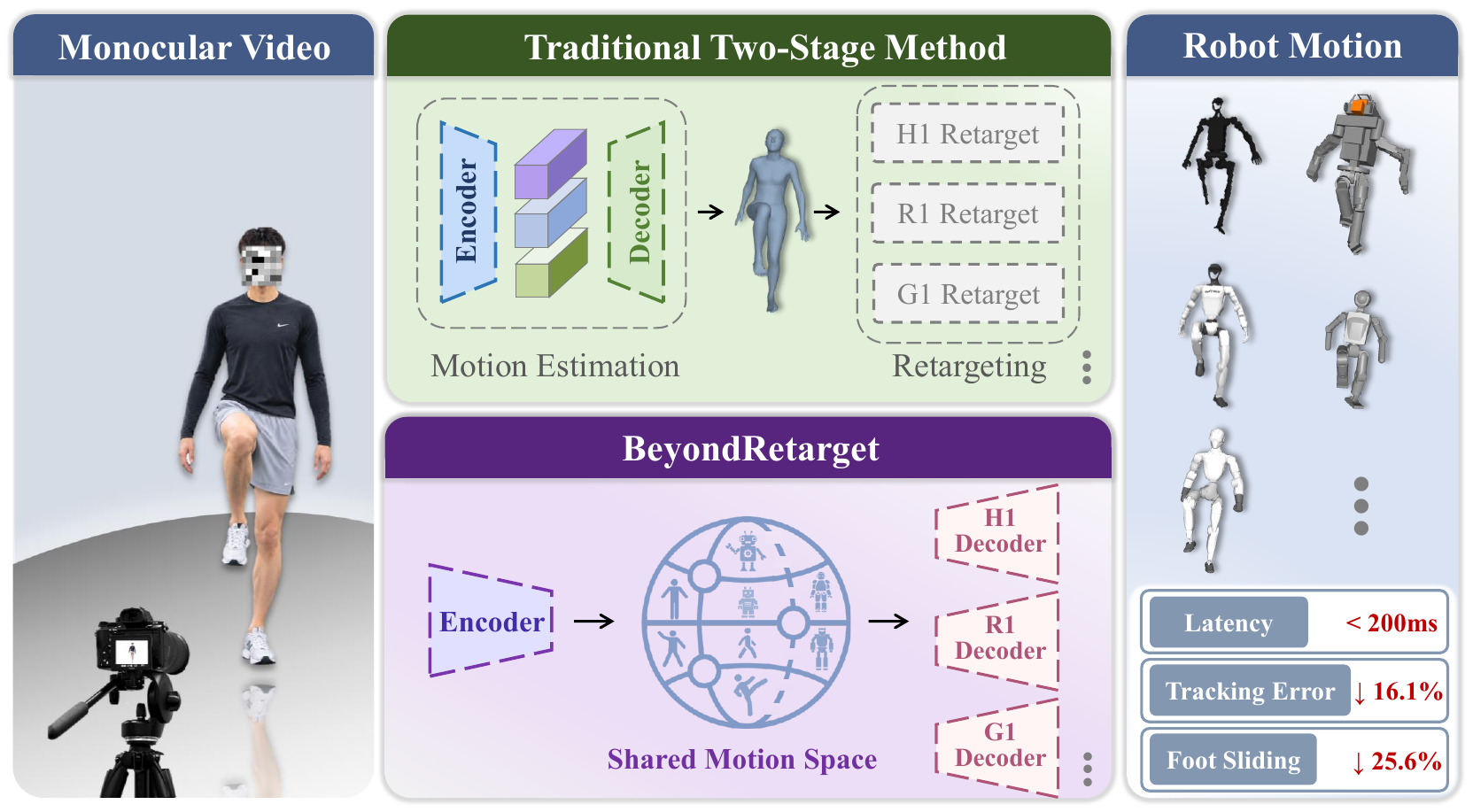}
  \caption{\method{} encodes the video into a shared motion space and directly decodes robot-specific trajectories for the requested humanoids.}
  \label{fig:pipeline}
\end{minipage}

\begin{abstract}
\input{sections/abstract.tex}
\end{abstract}

\input{sections/introduction.tex}
\input{sections/related_work.tex}
\input{sections/method.tex}
\input{sections/experiments.tex}

\input{sections/conclusion.tex}

\bibliography{references}
\bibliographystyle{iclr2027_conference}

\end{document}

%% file: math_commands.tex
\usepackage{amsmath,amsfonts,bm}

\def\eqref#1{equation~\ref{#1}}

\def\1{\bm{1}}

\DeclareMathAlphabet{\mathsfit}{\encodingdefault}{\sfdefault}{m}{sl}
\SetMathAlphabet{\mathsfit}{bold}{\encodingdefault}{\sfdefault}{bx}{n}

%% file: macros.tex
\newcommand{\method}{BeyondRetarget}
\newcommand{\rampjpe}{\textsc{MAMPJPE}}
\newcommand{\rte}{\textsc{RTE}}

\makeatletter
\newcommand{\wrapcaptionsetup}{\long\def\@makecaption##1##2{\vskip\abovecaptionskip
    {\scriptsize
      \sbox\@tempboxa{##1: ##2}\ifdim\wd\@tempboxa>\hsize
        ##1: ##2\par
      \else
        \global\@minipagefalse
        \hb@xt@\hsize{\hfil\box\@tempboxa\hfil}\fi
    }\vskip\belowcaptionskip
  }}
\makeatother

%% file: sections/abstract.tex
\vspace{-4mm}
Learning executable motions from human videos offers a scalable solution for humanoid robots to acquire demonstration motions. However, existing pipelines typically first construct an explicit human motion representation and then convert it into robot motions via motion retargeting. Although such methods can effectively leverage large volumes of existing human data for training, the substantial differences between humans and humanoid robots in locomotion mechanisms and joint degree-of-freedom configurations make motions generated by this human-representation-centric approach difficult to execute on robots. Furthermore, errors introduced during human motion estimation inevitably propagate to the retargeting stage and cannot be eliminated via joint optimization.
We propose \method{}, an end-to-end framework that directly maps monocular RGB videos to robot motions. Discarding the explicit human representation, this framework learns robot-oriented implicit representations directly from visual observations, enabling the model to capture cross-morphology motion structures. To generate motions more suitable for robot execution, we further design a contact-aware motion optimization mechanism to improve temporal consistency and physical plausibility. Experiments show that \method {} significantly improves the accuracy and robustness of generated robot motions, while achieving higher execution success rates and lower latency in both simulation environments and real humanoid robots.
Project page is available at
\href{https://bear-ty.github.io/Beyondretarget_page/}
{\textbf{\textcolor{blue}{BeyondRetarget}}}.
\vspace{-4mm}

%% file: sections/introduction.tex
\section{Introduction}
\vspace{-2mm}
Large-scale, executable motion data serve as a fundamental foundation for humanoid robots to perform skill learning and whole-body control. However, such data is typically acquired using specialized systems such as optical motion capture~\citep{vicon} and wearable inertial sensors~\citep{schepers2010ambulatory}, which require dedicated hardware and controlled capture environments, leading to high costs and difficulties in large-scale deployment~\citep{schepers2010ambulatory}. In contrast, monocular RGB videos are abundant and low-cost to collect. Therefore, monocular videos represent a highly promising data source for large-scale acquisition of robot motion data.

Benefiting from recent advances in human motion capture techniques, monocular RGB videos can be converted into unified explicit human motion representations~\citep{shin2024wham,shen2024gvhmr}, such as Skinned Multi-Person Linear Model (SMPL)~\citep{loper2015smpl}. Built upon this mature human motion representation, existing video-based robot motion generation methods typically adapt a two-stage pipeline: reconstructing human motion from RGB videos first, then retargeting human motion onto the target robot~\citep{fu2024humanplus,allshire2025visual,roux2026ddr}. This pipeline fully leverages the high similarity in overall structure and motion patterns between humans and humanoid robots, enabling extraction and transfer of meaningful motion patterns from human movements, and thus achieves promising results in most scenarios.

However, the similarity between humans and humanoid robots mainly lies in global morphology and kinematic structure, while notable discrepancies remain in body proportions, joint degrees of freedom (DoF), motion dynamics and other aspects. Consequently, the human-centric SMPL intermediate representation is not inherently suited for describing robot motion. Furthermore, the two-stage decomposition propagates human motion reconstruction errors from the first stage directly into the subsequent retargeting process, which compound with retargeting errors and form error accumulation that cannot be eliminated via joint optimization. Although the two-stage method provides a natural and effective solution for video-to-robot motion generation, its human-centric intermediate representation limits the capability to directly model motion within the robot space.


To address these issues, we propose \method{}, an end-to-end framework that directly generates executable humanoid robot motion from monocular RGB videos. The core idea is to eliminate the explicit SMPL intermediate representation and instead learn an implicit, robot-oriented motion representation directly from visual observations. This representation can be optimized directly for cross-morphology target robot motion generation. Furthermore, it can be mapped to motions of any target humanoid robot via robot-specific decoders. We establish geometric correspondences between humans and robots via a set of semantic keypoints and semantic link directions, and constructs a unified supervision mechanism to directly supervise robot motion using SMPL-based data. To improve motion executability, a contact-aware optimization module identifies foot contact states, determines support states, and performs kinematic corrections to enhance temporal consistency and physical plausibility. Figure~\ref{fig:pipeline} illustrates the core modifications brought by this work.

Experiments demonstrate that \method{} can generate accurate, robust, and executable humanoid robot motions from monocular videos. 
Compared with conventional two-stage pipelines, \method{} achieves lower pose errors and substantially reduces severe motion collapse, while more faithfully reproducing human motions and maintaining stable execution in both simulation and real-robot settings. Moreover, the shared motion features and unified supervision scheme enable efficient adaptation to diverse humanoid embodiments. These results demonstrate the potential of \method{} to transform the abundant and easily accessible monocular videos into high-quality executable robot motion data, providing a scalable and low-cost approach for large-scale robot motion data acquisition.

Our main contributions are:
\vspace{-2mm}
\begin{itemize}[leftmargin=2.0em]
    \item 
    We propose an end-to-end framework that \textbf{directly generates executable humanoid robot motion from monocular human videos}, providing a convenient solution for acquiring a large-scale data for humanoid robots.

    \item
    We propose an \textbf{implicit, shared robot-oriented motion representation}, together with a unified human-robot supervision mechanism, enabling this shared representation to be decoded onto different humanoid robot embodiments.

    \item 
    We introduce explicit contact and support representation to refine the estimated motion for improved temporal and \textbf{physical consistency}.

    \item
    We systematically \textbf{evaluate our method in both simulation and real-world robot platforms}, showing superior motion accuracy, robustness, execution success rate and real-time performance.
\end{itemize}

%% file: sections/related_work.tex
\vspace{-5mm}
\section{Related Work}
\vspace{-2mm}
Existing methods for generating robot motion data typically follow a two-stage pipeline~\citep{fu2024humanplus,allshire2025visual,roux2026ddr}, first recovering explicit human motion and then retargeting it to robots. This paper reviews related work from two aspects: human motion data acquisition and representation, and human-to-robot motion retargeting.

\textbf{Human motion acquisition and representation.}
Traditional systems capture human motion by tracking optical markers~\citep{vicon} or fusing wearable inertial and magnetic measurements~\citep{schepers2010ambulatory}. The captured motion is represented as marker trajectories~\citep{loper2014mosh} or skeletal joint positions and rotations~\citep{ionescu2014human36m,huang2018dip}. These systems provide detailed measurements but require dedicated hardware and venues, limiting collection at scale.
Parametric body models subsequently link skeletal motion to body shape and surface deformation~\citep{anguelov2005scape}. MoSh fits such models to sparse markers~\citep{loper2014mosh}, while SMPL provides a shared mesh topology and skeletal structure~\citep{loper2015smpl}. SMPL's consistent pose and shape parameters offer a common interface for learning and retargeting across heterogeneous data sources. MANO and SMPL-X extend this representation to articulated hands and expressive bodies~\citep{romero2017mano,pavlakos2019smplx}. AMASS unifies 15 optical motion-capture datasets in a common SMPL-based representation~\citep{mahmood2019amass}, while BEDLAM uses synthetic bodies and motion to scale image-based supervision~\citep{black2023bedlam}. MotionPRO and RICH further connect motion to pressure and scene contact~\citep{ren2025motionpro, lu2026pressmimic, huang2022rich}.

With these parametric body models providing a consistent representation of human motion, monocular RGB videos have emerged as an attractive source for scalable motion acquisition due to their abundance and ease of collection without specialized hardware. Single-image estimation evolved from model fitting to learned regression~\citep{bogo2016smplify,kanazawa2018hmr,goel2023humans4d}. Recovering motion, however, requires consistency across frames. Video methods therefore introduced temporal encoding and learned motion priors~\citep{kanazawa2019hmmr,kocabas2020vibe,luo2021meva}, followed by stronger temporal context and attention mechanisms~\citep{choi2021tcmr,wan2021maed}. 
WHAM further refines trajectories using contact information~\citep{shin2024wham}, while GVHMR introduces gravity-view coordinates to reduce global recovery ambiguity~\citep{shen2024gvhmr}. Together, these developments expand the data sources for motion acquisition, enabling monocular videos to be converted into structured human motion data, which can be transformed into corresponding robot action data through a retargeting pipeline.

\textbf{Human-to-robot motion retargeting.}
Retargeting transfers human motion to robots with different kinematics while preserving motion characteristics under robot constraints. 
Early approaches establish \textbf{hand-designed} correspondences between human and robot joints, followed by motion scaling, inverse kinematics,  and trajectory modification~\citep{pollard2002adapting,nakaoka2003dance,ude2004programming}. These methods establish the basic human-to-robot motion generation pipeline but require substantial robot-specific design. 
More recently, \textbf{optimization-based} methods formulate retargeting as constrained motion matching, jointly considering motion fidelity, morphology, and feasibility~\citep{liarokapis2013directions,ayusawa2017retarget,darvish2019retargeting}. GMR~\citep{joao2025gmr} is a representative recent approach that further improves motion fidelity and addresses physical artifacts. However, such non-convex optimization can be sensitive to initialization and objective design, leading to local collapse or discontinuous joint trajectories~\citep{zhao2026nmr}. 
To avoid these numerical instabilities, \textbf{learning-based} methods directly learn mappings from human to robot motion~\citep{villegas2018nkn,aberman2020skeleton}. NMR~\citep{zhao2026nmr}, for example, learns a robot motion mapping from high-quality paired supervision, improving inference stability and smoothness. However, the learned mapping is coupled to the target robot and training distribution, making adaptation to new robot morphologies costly. Another line of work~\citep{luo2026sonic,li2025bfmzero} maps human motion representations to latent commands for direct use in downstream control. While such latent interfaces reduce reliance on explicit retargeting, they provide less interpretable robot motion and limited access to complete trajectories for inspection and editing.

Overall, explicit robot motion remain a high-quality and indispensable source of data for robot motion learning and imitation. However, existing methods for acquiring such data typically decouple human motion capture from robot motion retargeting, resulting in inherent limitations such as information bottlenecks and error accumulation that  limit the quality of generated motion.

%% file: sections/method.tex
\vspace{-2mm}
\section{Method}
\vspace{-2mm}
\subsection{Problem Formulation and Framework Overview}
\label{sec:problem}
\vspace{-2mm}
Given a monocular RGB video sequence $\mathcal{I}=\{\mathbf{i}_t\}_{t=1}^{n}$ of length $n$ and a target robot configuration $r$, our goal is to directly estimate the complete motion of the target robot:
\begin{equation}
\hat{\mathcal{M}}^{(r)}
=
\{\hat{\mathbf{p}}_t,
\hat{\mathbf{o}}^{(r)}_t,
\hat{\mathbf{q}}^{(r)}_t\}_{t=1}^{n},
\label{eq:state}
\end{equation}
where $\hat{\mathbf{p}}_t \in \mathbb{R}^3$ denotes the root translation, while $\hat{\mathbf{o}}^{(r)}_t \in \mathbb{R}^6$ and $\hat{\mathbf{q}}^{(r)}_t \in \mathbb{R}^{d_r}$ denote the root orientation and the $d_r$-dimensional DoF vector of target robot $r$, respectively.

\begin{figure*}[t]
  \centering
  \includegraphics[width=\textwidth]{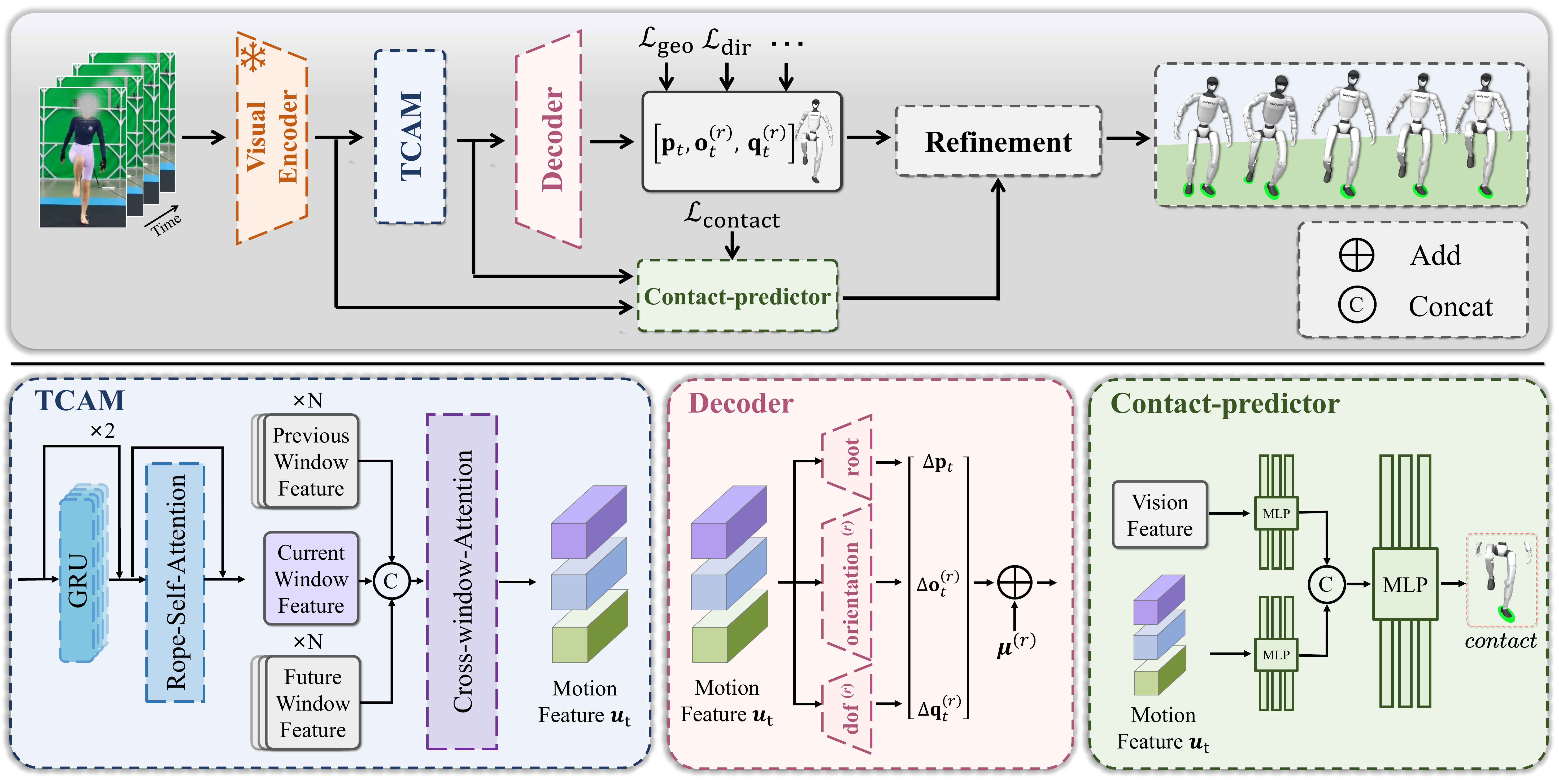}
  \caption{Model architecture. Input RGB video is encoded by the visual and
  temporal encoders, and the resulting shared motion representation is
  directly decoded into the target robot motion. Contact-aware refinement
  further improves support and temporal consistency.}
  \label{fig:model}
  \vspace{-5mm}
\end{figure*}

As illustrated in Figure~\ref{fig:model}, we use a frozen visual encoder~\citep{goel2023humans4d} to encode the video into visual features, which are then temporally modeled by the Temporal Context Aggregation Module (TCAM)~\citep{lu2026pressmimic} to produce a shared motion representation $\mathbf{u}_t$. This representation bridges the motion spaces of human and humanoid and can serve as a shared input for motion decoders of different robots.

For a target robot $r$, the robot motion decoder $\Psi_r$ directly maps the
shared motion representation to robot motion. We parameterize the motion as a
residual around a robot-specific nominal state:
\begin{equation}
[\hat{\mathbf{p}}_t,
\hat{\mathbf{o}}_t^{(r)},
\hat{\mathbf{q}}^{(r)}_t]
=
\boldsymbol{\mu}^{(r)}
+
[\Delta\mathbf{p}_t,\Delta\mathbf{o}_t^{(r)},\Delta\mathbf{q}_t^{(r)}],
\label{eq:direct-model}
\end{equation}
where
$[\Delta\mathbf{p}_t,\Delta\mathbf{o}_t^{(r)},\Delta\mathbf{q}_t^{(r)}]
=\Psi_r(\mathbf{u}_t)$ denotes the residuals for root translation, root orientation, and robot-specific joint motion, respectively, and
$\boldsymbol{\mu}^{(r)}\in \mathbb{R}^{3+6+d_r}$ is a learnable nominal state of robot $r$. 

\begin{wrapfigure}{R}{0.58\textwidth}
  \centering
  \includegraphics[width=\linewidth]{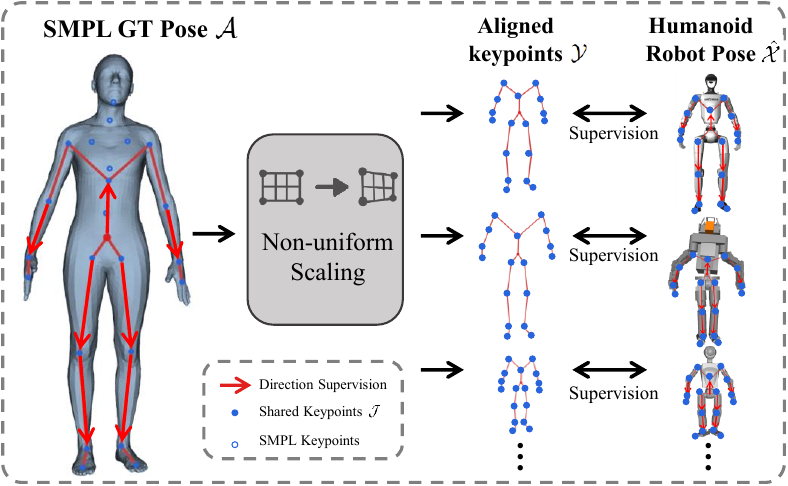}
  \vspace{-3mm}
  \caption{Unified human and robot supervision. Human keypoints are adapted
  to robot morphology. Keypoint positions and directed segments supervise humanoid robot pose.}
  \label{fig:space}
  \vspace{-7mm}
\end{wrapfigure}
The decoder consists of an MLP trunk, a root translation branch, a
root orientation branch, and a robot-specific Dof branch. For each additional
robot, the pretrained visual temporal representation, MLP trunk, and
root translation branch are reused, while its Dof and root orientation
branch are adapted to the robot's kinematic structure and degrees of freedom.

\vspace{-2mm}
\subsection{Unified Motion Supervision and Training}
\label{sec:space}
\vspace{-2mm}
Human pose parameters and robot's DoF differ in dimension and kinematic meaning, making supervision infeasible during training. In order to directly use human body annotations for training, we introduce a non-uniform scaling method that establishes a shared set of keypoints between human and robot, and adapts SMPL to the morphology of the target humanoid robot, thereby transferring motion supervision to a unified geometric space.

\textbf{Constructing the unified motion supervision.}
Given that all humanoid robots and the human body have similar topological structures, we define a set $\mathcal{J}$ of 18 shared keypoints covering the pelvis, torso, shoulders, elbows, wrists, hands, hips, knees, ankles, and feet. On the human side, these keypoints are extracted from the SMPL joints, denoted as $\mathcal{A} = \{\mathbf{a}_{t,j}\in\mathbb{R}^{3}\}$ where $t\in\{1,\ldots,n\}$ is the frame index and $j\in\mathcal{J}$ is the keypoint index. On the robot side, keypoints $\mathcal{\hat{X}} = \{\hat{\mathbf{x}}_{t,j}\in\mathbb{R}^{3}\}$ are a series of corresponding joints. 

To align the morphology of the human and the robot, we scale each link of the skeleton independently. Let $\pi(j)$ denote the parent of keypoint $j$, and let $\ell_j^{(\mathrm{h})}$ and $\ell_j^{(\mathrm{r})}$ denote the lengths of link $(\pi(j),j)$ for the human reference and the target robot, respectively. Taking the pelvis keypoint $0$ as the root, we construct the robot-proportioned target as
\begin{equation}
\mathbf{y}_{t,0} = \mathbf{a}_{t,0},\qquad
\mathbf{y}_{t,j}
=
\mathbf{y}_{t,\pi(j)}
+
\frac{\ell_j^{(\mathrm{r})}}{\ell_j^{(\mathrm{h})}}
\left(\mathbf{a}_{t,j} - \mathbf{a}_{t,\pi(j)}\right).
\label{eq:morphology}
\vspace{-3mm}
\end{equation}

The resulting $\mathbf{y}_{t,j}$ forms the robot-proportioned supervision target. Human annotations can therefore directly supervise robot motions without native robot trajectories.

For each frame, we align the estimated robot keypoint set $\mathcal{\hat{X}}$ to the scaled human keypoint set $\mathcal{Y}=\{\mathbf{y}_{t,j}\in\mathbb{R}^{3}\}$ via the Procrustes transformation $\phi_t$.
Let $\tilde{\mathbf{x}}_{t,j} = \phi_t(\hat{\mathbf{x}}_{t,j},\mathbf{y}_{t,j})$ denote the robot keypoint after alignment.
The geometry loss is
\begin{equation}
\mathcal{L}_{\mathrm{geo}}
=
\frac{1}{n|\mathcal{J}|}
\sum_{t=1}^{n}
\sum_{j\in\mathcal{J}}
\rho\!\left(
\tilde{\mathbf{x}}_{t,j}
- \mathbf{y}_{t,j}
\right),
\label{eq:geometry}
\end{equation}
where $\rho$ denotes the coordinate-wise Smooth $L_1$ penalty and $|\cdot|$ denotes the cardinality of a set. 

Keypoint positions do not fully constrain local orientation, so we define a set $\mathcal{D}$ of directed semantic links corresponding to the upper arms, forearms, thighs, shanks, and torso. For $d\in\mathcal{D}$, 
let $\hat{\mathbf{v}}_{t,d}$ denote the normalized estimated link vectors computed by $\hat{\mathbf{x}}_{t,j}$, and $\mathbf{v}_{t,d}$ denote the target link vectors computed by $\mathbf{y}_{t,j}$, respectively. The direction loss penalizes their cosine discrepancy:
\begin{equation}
\mathcal{L}_{\mathrm{dir}}
=
\frac{1}{n|\mathcal{D}|}
\sum_{t=1}^{n}
\sum_{d\in\mathcal{D}}
\left(1-
\hat{\mathbf{v}}_{t,d}^{\!\top}
\mathbf{v}_{t,d}
\right).
\label{eq:direction}
\vspace{-1mm}
\end{equation}

This constraint can effectively improve the accuracy of the link direction and make the robot's motion more natural and realistic. 


\textbf{Training objective and multiple robots.}
The complete training loss combines geometry, direction, temporal consistency, and robot joint limits:
\begin{equation}
\mathcal{L}_{\mathrm{motion}}
=
\lambda_{\mathrm{geo}}\mathcal{L}_{\mathrm{geo}}
+ \lambda_{\mathrm{dir}}\mathcal{L}_{\mathrm{dir}}
+ \lambda_{\mathrm{trans}}\mathcal{L}_{\mathrm{trans}}
+ \lambda_{\mathrm{ori}}\mathcal{L}_{\mathrm{ori}}
+ \lambda_{\mathrm{temp}}\mathcal{L}_{\mathrm{temp}}
+ \lambda_{\mathrm{lim}}\mathcal{L}_{\mathrm{lim}}.
\label{eq:objective}
\vspace{-1mm}
\end{equation}

Each $\lambda_{\{\cdot\}}$ denotes the weight for its corresponding loss term.
$\mathcal{L}_{\mathrm{geo}}$ and $\mathcal{L}_{\mathrm{dir}}$ supervise the relative robot pose within the unified motion space.
$\mathcal{L}_{\mathrm{trans}}$ and $\mathcal{L}_{\mathrm{ori}}$ constrain the root translation and rotation motion, respectively.
$\mathcal{L}_{\mathrm{temp}}$ enforces temporal consistency across consecutive frames,
and $\mathcal{L}_{\mathrm{lim}}$ penalizes violations of robot joint limits.

We first train the encoder and complete robot motion decoder on Unitree G1 robot. For a new robot, we configure its kinematic tree, semantic keypoints, and morphology scales, reuse the encoder, TCAM and root translation branch in decoder, and only train its robot-specific Dof and root orientation branch.

\vspace{-2mm}
\subsection{Contact-Aware Refinement}
\label{sec:refinement}
\vspace{-2mm}
\begin{figure*}[t]
  \centering
  \includegraphics[width=\textwidth]{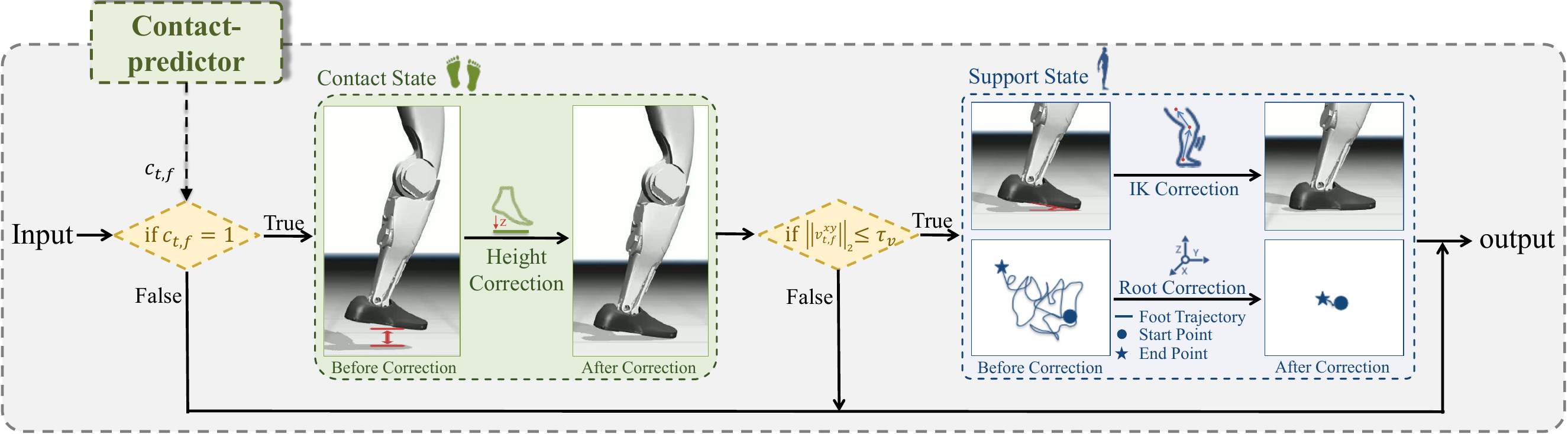}
  \caption{Contact-aware kinematic refinement. The robot's lower limb are divided into contact or support states based on explicit contact representation, and optimized via corresponding post-processing to improve the physical plausibility.}
  \label{fig:refinement}
  \vspace{-5mm}
\end{figure*}

The estimator is capable of generating complete robot motion. However, accurate poses inferred from visual inputs do not necessarily yield physically executable motions. This is because the executability of robot motion largely depends on temporal consistency and contact stability, whereas the unified motion supervision scheme provides insufficient constraints on these two properties. To address this issue, we first freeze the motion estimator, filter its generated motions, then introduce explicit contact representations to guide subsequent motion optimization.

Specifically, after motion training, we fix estimator's parameters and train a contact predictor under pressure data supervision~\citep{ren2025motionpro}. This predictor fuses the motion feature with the visual feature to predict the contact probability of each foot. As illustrated in Figure~\ref{fig:refinement}, we use a two-level state decision. First, we convert the predicted contact probability $\hat{c}_{t,f}\in[0,1]$ of each foot $f$ into a binary contact state $c_{t,f}\in\{0,1\}$, where $0$ denotes no contact and $1$ denotes contact. We then obtain the foot trajectory and its horizontal velocity $v_{t,f}^{xy}$ through forward kinematics. A contacted foot is regarded as being in the support state only when its horizontal velocity is below the threshold $\tau_v$:
\begin{equation}
s_{t,f}
=
c_{t,f}
\land
\left(
\left\|v_{t,f}^{xy}\right\|_2\leq\tau_v
\right).
\label{eq:support}
\vspace{-1mm}
\end{equation}
$||\cdot||_2$ denotes the vector norm. This distinction separates feet that are in contact with the ground but still moving from those providing stable support. The contact state triggers root height correction: we eliminate height offsets to prevent foot penetration or floating. The support state additionally enables root horizontal correction and lower-body inverse kinematics (IK) solving to ensure no sliding of stably supporting feet. This hierarchical structure maintains stable support feet while naturally allowing sliding for moving contact feet. To avoid abrupt changes caused by support state switching, we apply a temporal smoothing filter $\operatorname{Filt}(\cdot)$ to the residual sequence and add the filtered correction to the original prediction to obtain the final refined joint posture. Let $\mathbf{\hat{q}}_{t}$ denote the raw predicted full-body joint posture at frame $t$, and $\mathbf{q}_{t}^{\mathrm{IK}}$ denote the optimized joint posture solved via constrained inverse kinematics. The IK correction residual across all frames $t\in\{1,\ldots,n\}$ is formulated as:
\begin{equation}
\Delta\mathbf{q}_{t}^{\mathrm{IK}}
=
\mathbf{q}_{t}^{\mathrm{IK}}
-
\mathbf{\hat{q}}_{t},\qquad
\mathbf{\hat{q}}_{t}
\leftarrow{}
\mathbf{\hat{q}}_{t}
+
\operatorname{Filt}
\left(
\Delta\mathbf{q}_{t}^{\mathrm{IK}}
\right).
\label{eq:ik_filter}
\end{equation}
The filtered correction ensures smooth and physically consistent motion transitions while maintaining accurate foot contact and support constraints.

%% file: sections/experiments.tex
\vspace{-2mm}
\section{Experiments}
\label{sec:experiments}
\vspace{-2mm}
We organize the experiments around four questions: (1) Can our end-to-end method outperform the two-stage methods that first reconstruct human motion and then retarget it; 
(2) Can shared motion representation adapt to multiple humanoid robots; 
(3) How does each component affect model performance; 
(4) How does the motion output by our method perform on a real robot? Can it support real-time motion control?

\vspace{-2mm}
\subsection{Experimental setup}
\vspace{-2mm}
\textbf{Setup.}
We train on MotionPRO~\citep{ren2025motionpro}, which provides synchronized RGB images, optical motion capture SMPL ground truth (GT), and foot contact labels, covering approximately 12,000,000 frames and over 400 action categories. SMPL GT supervises robot motion, while foot contact labels are used to train the contact predictor. We train the model on Unitree G1, and adapt each additional humanoid (Unitree R1, Fourier GR1-T1, Fourier GR2-V3, Unitree H1, Booster T1, Tienkung, and Atlas) by retraining its joint and root orientation branches in the robot‑specific decoder. Adapting a new robot takes approximately one hour on a single RTX~4090. We compare against GVHMR~\citep{shen2024gvhmr}/WHAM~\citep{shin2024wham} combined with NMR~\citep{zhao2026nmr}/GMR~\citep{joao2025gmr}. To isolate errors from human motion reconstruction, Oracle baseline uses SMPL GT for direct retargeting.

\textbf{Metrics.}
We evaluate local pose, global trajectory, temporal continuity, physical plausibility and executability through the following metrics. Morphology-Aligned MPJPE (\rampjpe{}, mm) measures the keypoint error after aligning the reference human body with the robot morphology, while Root Trajectory Error (\rte{}, mm) measures root trajectory error after rigid alignment. Torso Direction Error (TDE, $^\circ$) and End-effector Direction Error (EDE, $^\circ$) are used to measure the angular errors of the torso vector and the end-effector vector relative to the GT vector, respectively. Jitter (10 m/s$^3$) and Accel (mm/frame$^2$) measure higher order temporal jitter and acceleration error. Foot sliding (FS, mm/frame) measures the unintended horizontal displacement of the feet during support frames, while Violation rate (Viol., \%) measures the proportion of contact frames exhibiting foot floating or ground penetration. Fail$_{65}$ (\%) and Fail$_{100}$ (\%) measure the proportions of windows whose average \rampjpe{} exceeds 65~mm and 100~mm, respectively, over the total number of windows (each consists of 200 frames). Executed Success Rate (Exec. SR, \%) measures the ratio of successfully executed clips to the total number of clips in physical simulation (Mujoco). We use a general motion tracker SONIC~\citep{luo2026sonic} to execute the output motion, and a motion clip (each consists of 600 frames) is considered failed if its execution is terminated before reaching the end of the reference motion due to excessive root height, root orientation, or end-effector height errors. Executed Morphology-Aligned MPJPE (Exec. \rampjpe{}, mm) measures the accuracy of the executed motion.

\begin{table*}[t]
\caption{Quantitative comparison on G1. The oracle rows at the top use SMPL GT to isolate errors from human motion reconstruction, and thus are excluded from the ranking.}
\vspace{-2mm}
\label{tab:main}
\centering
\scriptsize
\resizebox{\textwidth}{!}{\begin{tabular}{lccccccccc}
\toprule
Method & \rampjpe{} $\downarrow$ & \rte{} $\downarrow$ & Jitter $\downarrow$ & Accel $\downarrow$ &
Fail$_{65}$ $\downarrow$ & Fail$_{100}$ $\downarrow$ & FS $\downarrow$ &
Exec. SR $\uparrow$ & Exec. \rampjpe{} $\downarrow$ \\
\midrule
GT$\rightarrow$NMR & 33.26 & 837.62 & 1.71 & 1.16 & 3.99 & 1.39 & 1.30 & 346/362 & 40.32 \\
GT$\rightarrow$GMR & 32.02 & 36.01 & 10.65 & 2.27 & 0.00 & 0.00 & 0.82 & 347/362 & 35.93 \\
\midrule
GVHMR$\rightarrow$NMR & \cellcolor{Second}39.19 & 701.13 & \cellcolor{Second}1.63 & \cellcolor{Second}1.56 & 6.21 & 2.22 & \cellcolor{Second}2.28 & 340/362 & 43.32 \\
GVHMR$\rightarrow$GMR & 39.46 & 637.33 & 7.94 & 2.51 & \cellcolor{Second}3.43 & \cellcolor{Second}0.28 & 3.30 & 313/362 & \cellcolor{Second}41.77 \\
WHAM$\rightarrow$NMR & 43.15 & 849.36 & 4.87 & 3.07 & 9.36 & 4.17 & 6.64 & \cellcolor{Second}342/362 & 46.56 \\
WHAM$\rightarrow$GMR & 45.88 & \cellcolor{Second}217.76 & 63.26 & 14.57 & 10.47 & 6.30 & 13.30 & 295/362 & 42.89 \\
\midrule
\textbf{\method{}} & \cellcolor{First}26.36 & \cellcolor{First}80.74 & \cellcolor{First}1.41 & \cellcolor{First}1.51 & \cellcolor{First}0.00 & \cellcolor{First}0.00 & \cellcolor{First}1.71 & \cellcolor{First}347/362 & \cellcolor{First}35.04 \\
\bottomrule
\end{tabular}}
\vspace{-2mm}
\end{table*}

\vspace{-2mm}
\subsection{Comparison with Two-Stage Methods}
\vspace{-2mm}
\begin{figure*}[t]
\centering
\includegraphics[width=\textwidth]{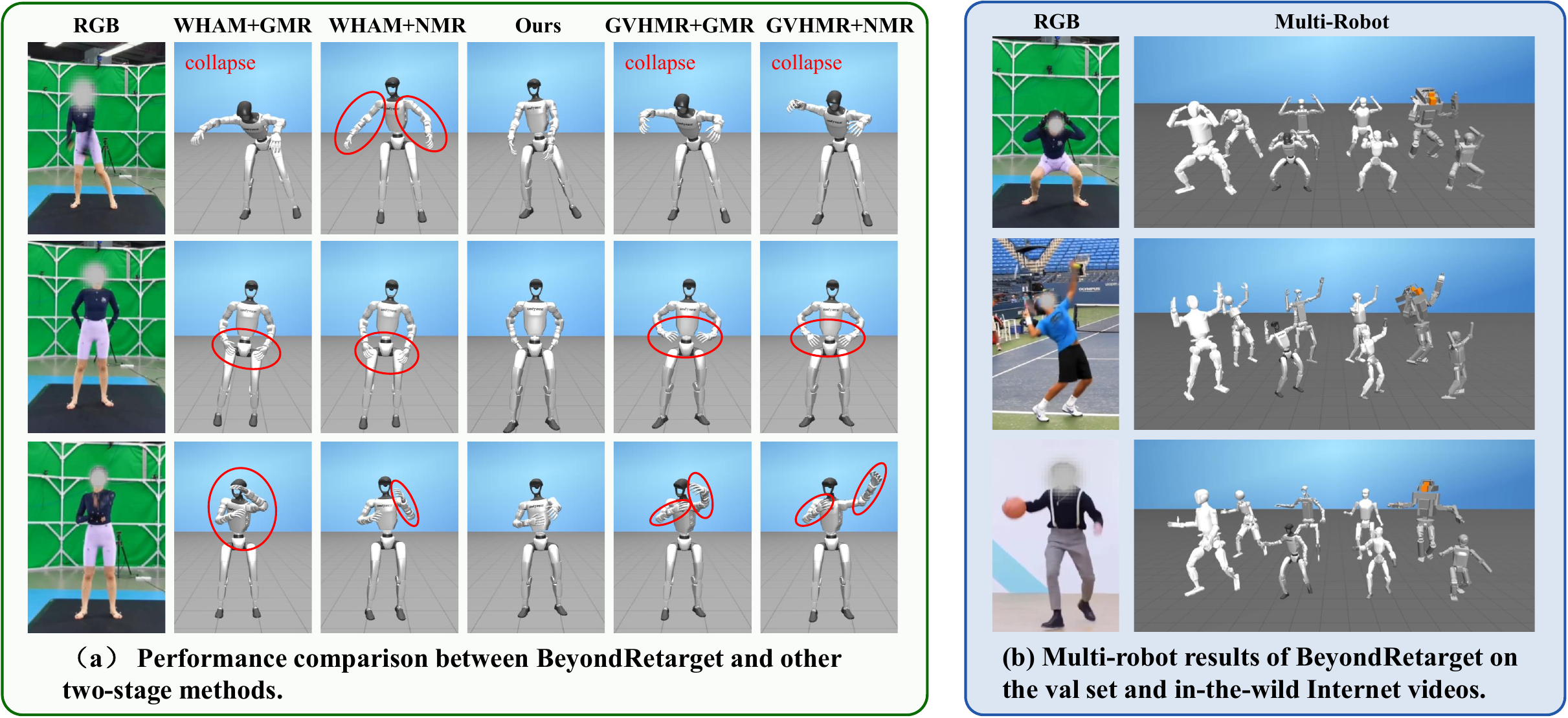}
\caption{Qualitative results. }
\label{fig:qualitative}
\vspace{-6mm}

\end{figure*}

Table~\ref{tab:main} compares pose, trajectory, temporal consistency, and physical plausibility. \method{} consistently outperforms all two-stage methods across \rampjpe{}, RTE, Jitter, and Foot Slide, demonstrating that end-to-end estimation improves motion accuracy, temporal consistency and physical plausibility. More importantly, \method{} eliminates severe pose collapse: no evaluated window exceeds either the 65~mm or 100~mm \rampjpe{} threshold, while all two-stage methods exhibit such collapse. This indicates that removing the intermediate human representation not only improves average accuracy but also substantially enhances robustness.
As illustrated in Figure~\ref{fig:qualitative}(a), for large-amplitude motions such as arm swings, two-stage pipelines suffer from misplaced limbs and pose collapse, whereas the direct estimation approach preserves motion characteristics more faithfully. For fine-grained actions, such as the hands-behind-the-waist poses, all two-stage methods incorrectly place the hands in front of the body, whereas \method{} accurately reconstructs the hands-behind-the-back motion.

To isolate retargeting error from human estimation error, we evaluate an Oracle setting where SMPL GT is directly fed into the retargeting module. \method{} still achieves lower \rampjpe{} than both Oracle retargeting baselines, demonstrating that the retargeting stage itself introduces non-negligible motion errors and providing further evidence for avoiding the intermediate human model.

Finally, in physical simulation, \method{} achieves a 347/362 (95.86\%) success rate with a Exec. \rampjpe{} of 35.04~mm, outperforming all two-stage pipelines using estimated human motion. Together, these results show that direct estimation improves RGB-to-robot motion accuracy, avoids pose collapse introduced by retargeting, and produces more executable reference motions.

\vspace{-4mm}
\subsection{Multi-robot experiments}
\vspace{-2mm}
\begin{table*}[t]
\caption{Multi-robot \rampjpe{} evaluation. The ``--'' entries denote robots unsupported by GMR.}
\vspace{-1mm}
\label{tab:multi}
\centering
\scriptsize
\resizebox{\textwidth}{!}{\begin{tabular}{lcccccccc}
\toprule
Method & H1 & T1 & Tienkung & G1 & R1 & GR1-T1 & GR2-V3 & Atlas \\
\midrule
GVHMR$\rightarrow$GMR & 77.79 & 70.09 & 103.23 & 39.46 & -- & -- & -- & -- \\
WHAM$\rightarrow$GMR & 99.66 & 77.26 & 115.86 & 45.88 & -- & -- & -- & -- \\
\midrule
\textbf{\method{}}& \textbf{51.32} & \textbf{51.25} & \textbf{67.99} & \textbf{26.36} & \textbf{27.81} & \textbf{43.78} & \textbf{49.06} & \textbf{72.05} \\
\bottomrule
\end{tabular}}
\vspace{-3mm}
\end{table*}
Table~\ref{tab:multi} evaluates the same adaptation procedure on eight humanoids with different morphologies. NMR is not included because it currently supports only the Unitree G1 and adapting it to a new robot morphology requires corresponding robot data. Due to differences in joint structures, different robots have varying \rampjpe{} values, but all falling within a reasonable range of 26.36~mm to 72.05~mm, and  our method outperforms GVHMR$\rightarrow$GMR on all four robots supported by both methods. Figure~\ref{fig:qualitative}(b) further shows that all eight humanoids can reproduce the same visual action, with performance generalizing to in‑the‑wild Internet videos, which indicates that the unified supervision can transfer the shared visual motion representation across different robot embodiments and possesses strong visual generalization capability.

\vspace{-4mm}
\subsection{Ablation studies}
\vspace{-2mm}
Table~\ref{tab:ablation} presents the metric results obtained by separately ablating non-uniform morphology alignment, $\mathcal{L}_{\mathrm{dir}}$, $\mathcal{L}_{\mathrm{temp}}$, and contact-aware refinement. First, removing non-uniform morphology alignment increases \rampjpe{} from 26.36~mm to 36.08~mm, the largest pose degradation among the ablations, demonstrating non-uniform morphology alignment can provide more reasonable supervision. Removing $\mathcal{L}_{\mathrm{dir}}$ increases \rampjpe{}, TDE and EDE, demonstrating that explicitly constraining semantic link directions enables more accurate local orientations of body parts.

\begin{table*}[t]
\caption{Component ablation results.}
\vspace{-2mm}
\label{tab:ablation}
\centering
\scriptsize
\resizebox{\textwidth}{!}{\begin{tabular}{lcccccccc}
\toprule
Variant & \rampjpe{} $\downarrow$ & \rte{} $\downarrow$ & TDE $\downarrow$ & EDE $\downarrow$ & Jitter $\downarrow$ & Accel $\downarrow$ & FS $\downarrow$ & Viol. $\downarrow$ \\
\midrule
w/o non-uniform alignment & 36.08 & 89.84 & 5.17 & 15.77 & 1.60 & 1.58 & 1.89 & 4.39 \\
w/o $\mathcal{L}_{\mathrm{dir}}$ & 29.62 & 94.77 & 6.95 & 21.78 & 1.45 & 1.59 & 2.08 & 7.00 \\
w/o $\mathcal{L}_{\mathrm{temp}}$ & 26.44 & 77.12 & 2.95 & 15.41 & 1.93 & 1.61 & 1.99 & 5.66 \\
w/o refinement & 25.73 & 91.33 & 2.95 & 14.31 & 2.17 & 1.74 & 4.51 & 22.18 \\
\textbf{Full model} & 26.36 & 80.74 & 2.95 & 15.29 & 1.42 & 1.51 & 1.71 & 6.17 \\
\bottomrule
\end{tabular}}
\vspace{-5mm}
\end{table*}

Second, removing $\mathcal{L}_{\mathrm{temp}}$ substantially degrades Jitter, Accel, and FS, demonstrating its effectiveness in improving temporal continuity. Finally, removing contact-aware refinement substantially degrades \rte{}, FS, and Viol. This shows that refinement corrects the raw trajectories and improves physical plausibility while preserving action accuracy.

\vspace{-2mm}
\subsection{Real‑robot and teleoperation experiments}
\vspace{-2mm}
\begin{figure*}[t]
  \centering
  \includegraphics[width=\textwidth]{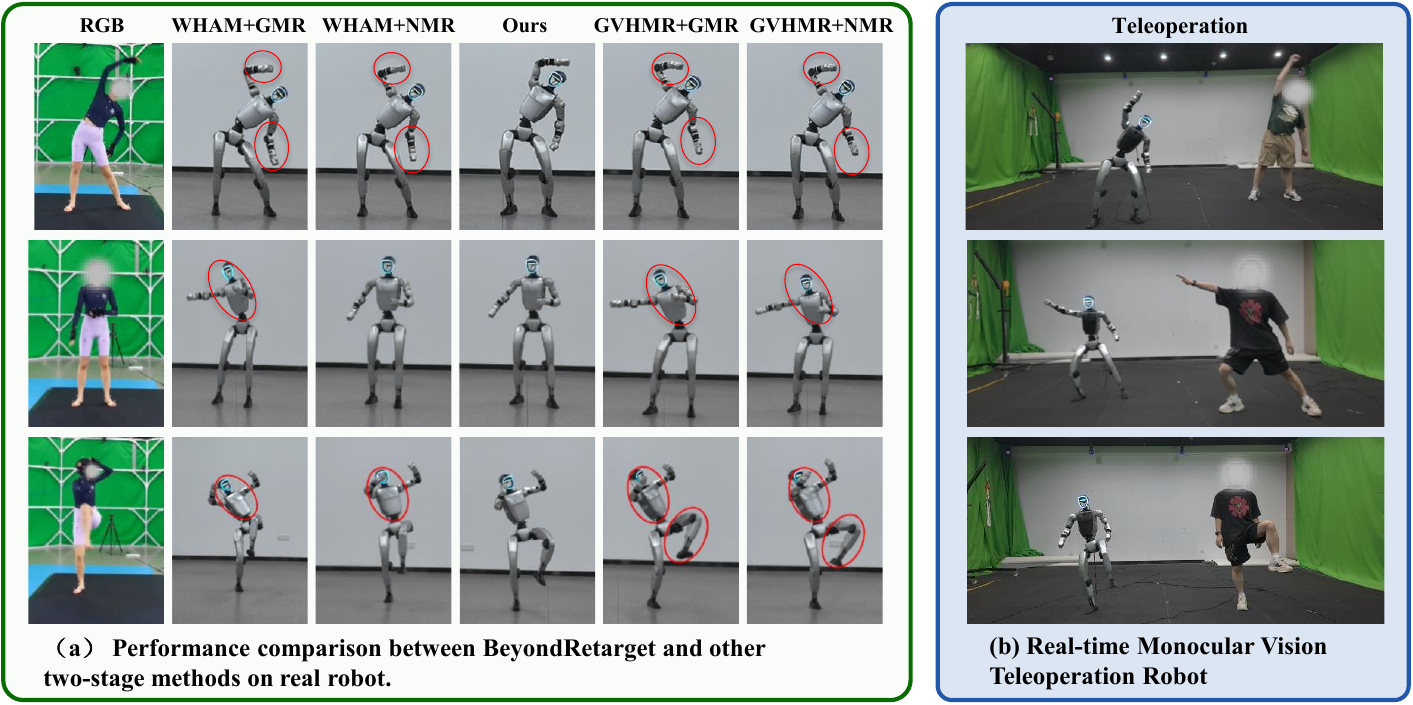}
  \caption{Real-robot evaluation and visual teleoperation. }
  \label{fig:realrobot}
  \vspace{-6mm}
\end{figure*}

Figure~\ref{fig:realrobot}(a) compares five methods executing the same input motions on a real robot. For the stretching motion, \method{} successfully extends the left arm above the head with the hand kept close to the thigh, whereas the four two-stage baselines exhibit clear arm motion discrepancies. For the single-arm circling and hands-on-head motion, \method{} maintains stable body motion and foot support throughout execution, while all four two-stage baselines show apparent instability. Together with the simulation results in Table~\ref{tab:main}, these results demonstrate that the improvements in motion accuracy and physical stability achieved in physical simulation transfer effectively to real-robot execution, further validating the superior executability of \method{}. 

\begin{wraptable}{R}{0.58\textwidth}
  \centering
  \vspace{-4mm}
  \caption{Real-time performance comparison.}
  \label{tab:streaming}
  \vspace{-2mm}
  \resizebox{\linewidth}{!}{
  \begin{tabular}{lccc}
  \toprule
  Method & \makecell{latency (ms)} & \makecell{throughput (FPS)} & \makecell{VRAM (GiB)} \\
  \midrule
  GVHMR$\rightarrow$GMR & 1369.4 & 44.31 & 9.18 \\
  GVHMR$\rightarrow$NMR & 1305.7 & 44.36 & 9.73 \\
  \midrule
  \textbf{\method{}} & \textbf{192.8} & \textbf{50.02} & \textbf{7.03} \\
  \bottomrule
  \end{tabular}
  }
  \vspace{-3mm}
\end{wraptable}

On the same GPU, we build streaming inference pipelines for different methods and configure each to achieve the lowest possible latency, then measure their median latency and peak VRAM usage. We separately test their maximum throughput without limiting the input frame rate.
Table~\ref{tab:streaming} shows that, benefiting from our end-to-end framework and streamlined model architecture, \method{} achieves lower output latency, reduced peak GPU memory usage, and higher maximum throughput compared to both two-stage baselines. This efficiency supports responsive real-time visual teleoperation, as demonstrated in Figure~\ref{fig:realrobot}(b).

%% file: sections/conclusion.tex
\vspace{-2mm}
\section{Conclusion and Limitations}
\vspace{-4mm}
\textbf{Conclusion.}
We present \method{}, an end-to-end framework that generates executable humanoid robot motion from monocular human videos. Experiments demonstrate that our framework achieves higher accuracy, robustness, executability and real-time performance compared with two‑stage pipelines. It can also scale to adapt to diverse humanoid embodiments, offering a new scheme for convenient, large‑scale acquisition of multi-robot motion data.

\textbf{Limitations.}
Constrained by model capacity and training data distribution, our method can still exhibit errors in global trajectory estimation for large-range human motions or when the camera is noticeably moving. In addition, our current method do not cover complex terrains or contact-rich human–environment interaction scenarios. Future work could incorporate more diverse data and model scene geometry and physical contacts, thus enabling robots to directly learn more challenging, scene-conditioned motion skills from videos.